\documentclass{article}

\PassOptionsToPackage{numbers}{natbib}
\usepackage[preprint]{defaultpk}
\usepackage{amsmath}   
\usepackage{hyperref}       
\usepackage{url}   
\usepackage{cleveref}  
\usepackage{graphicx}
\usepackage{algorithm}
\usepackage{algpseudocode}
\usepackage{array}    
\usepackage{xcolor}  
\usepackage[table]{xcolor}
\usepackage{makecell}

\usepackage[utf8]{inputenc} 
\usepackage[T1]{fontenc}             
\usepackage{booktabs}      
\usepackage{amsfonts}       
\usepackage{nicefrac}       
\usepackage{microtype}      
\usepackage{xcolor}         
\usepackage{hyperref}
\usepackage{multirow}
\usepackage{graphicx} 
\usepackage{amsmath}
\usepackage{enumitem}
\usepackage{subfigure}
\usepackage{comment}
\usepackage{marvosym}
\usepackage{caption}
\usepackage{wrapfig}
\usepackage{float}
\usepackage[normalem]{ulem}
\usepackage{amssymb}
\usepackage{wasysym}

\newcommand{\std}[1]{\textcolor{gray}{\scriptsize$\pm$#1}}
\title{Continually Evolving Skill Knowledge in \\
Vision Language Action Model }

\usepackage{pifont}

\makeatletter
\renewcommand{\@fnsymbol}[1]{%
  \ifcase#1\or
  *\or
  \mathsection\or
  \ddagger \or
  \dagger\or
  \mathsection\or
  \mathparagraph\or
  \|\or
  **\or
  \dagger\dagger\or
  \ddagger\ddagger
  \fi
}
\makeatother

\author{
    Yuxuan Wu$^{1,2}$,
    Guangming Wang$^{3}$,
    Zhiheng Yang$^{1}$,
    Tianchen Deng$^{1,4,5}$,
    \\
    \textbf{Maoqing Yao}$^{6}$,
    \textbf{Brian Sheil}$^{3}$,
    \textbf{Hesheng Wang}$^{1}$\thanks{\textbf{Corresponding author.} The authors affiliated with Shanghai Jiao Tong University (Yuxuan Wu, Zhiheng Yang, Tianchen Deng, and Hesheng Wang) are with the School of Automation and Intelligent Sensing, Shanghai Jiao Tong University, and the Shanghai Key Laboratory of Navigation and Location Based Services, Shanghai 200240, China.} \\
    \fontsize{9pt}{10pt}\selectfont{$^1$Shanghai Jiao Tong University}
    \quad
    \fontsize{9pt}{10pt}\selectfont{$^2$Shanghai Innovation Institute}
    \quad
    \fontsize{9pt}{10pt}\selectfont{$^3$University of Cambridge}
    \\
    \fontsize{9pt}{10pt}\selectfont{$^4$Nanyang Technological University}
    \quad
    \fontsize{9pt}{10pt}\selectfont{$^5$MIT SMART}
    \quad
    \fontsize{9pt}{10pt}\selectfont{$^6$AgiBot} \\
    \fontsize{9pt}{11pt}\selectfont
    \texttt{\{furrygreen, zhiheng.yang, wanghesheng\}@sjtu.edu.cn} 
    \texttt{\{gw462, bbs24\}@cam.ac.uk}
}

\begin{document}

\maketitle

\begin{figure}[h]
\centering
\includegraphics[width=0.85\textwidth]{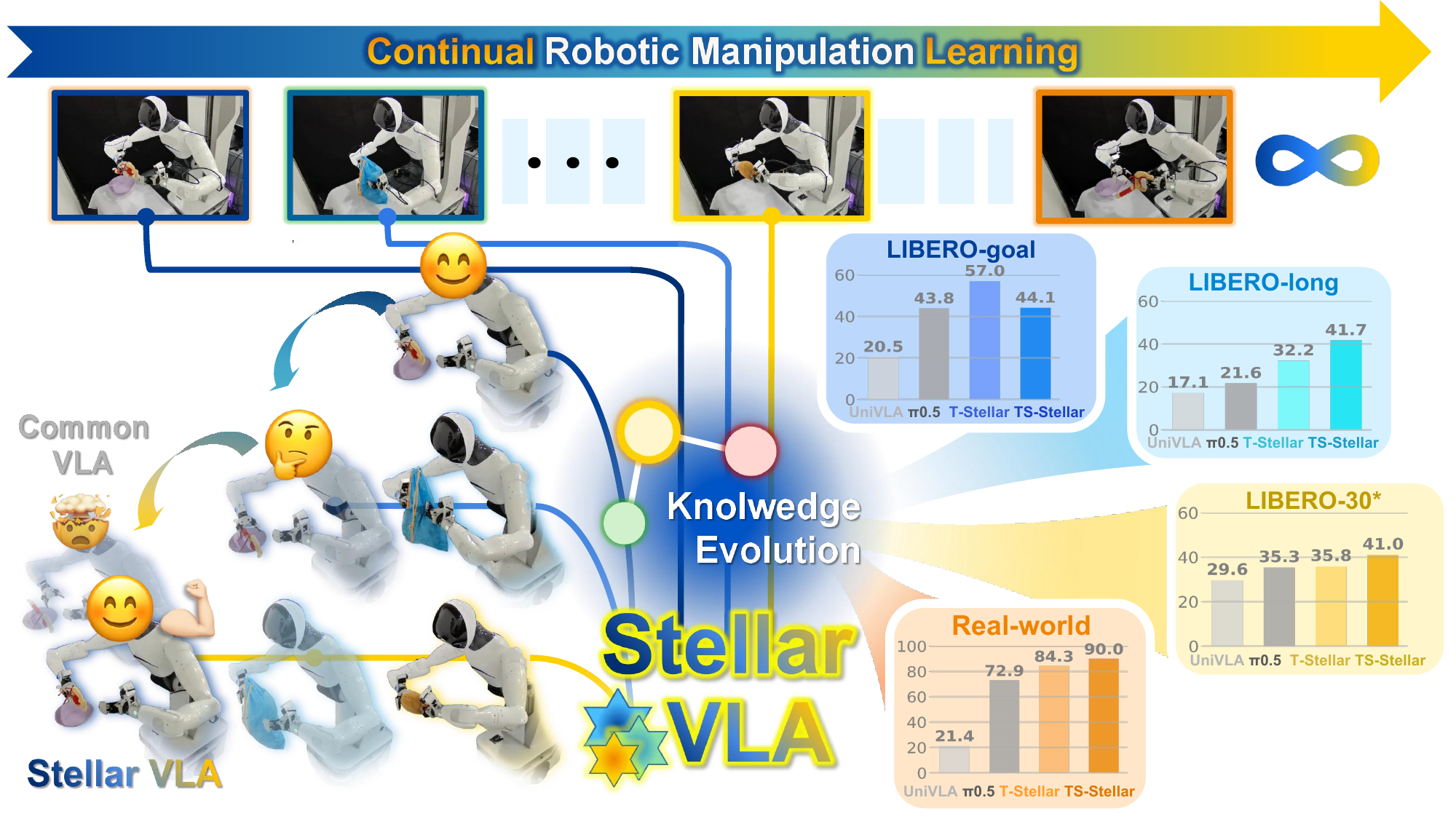}
\caption{Illustration of {Stellar VLA}. T-Stellar and TS-Stellar outperform UniVLA~\cite{UniVLA} and $\pi_{0.5}$~\cite{pi05} in final average success rates across simulation and real-world dual-arm experiments.}
\label{fig:teaser}
\end{figure}

\begin{abstract}
Vision-language-action (VLA) models show promising knowledge accumulation ability through pretraining, yet continual learning remains challenging. Existing continual imitation learning (CIL) methods often require additional parameters or modules, limiting scalability for large VLA models. We propose Stellar VLA, a knowledge-driven CIL framework without parameter growth, featuring two variants: T-Stellar for task-centric modeling and TS-Stellar for hierarchical modeling of reusable skills across tasks. Stellar VLA jointly optimizes task representations and a self-evolving knowledge space, with knowledge-posterior-guided expert routing using knowledge relations and Top-$K$ semantic embeddings for task specialization. Experiments on LIBERO demonstrate strong performance against VLA and CIL baselines with only 1\% replay. Dual-arm real-world experiments validate knowledge transfer across embodiments and scenes. TS-Stellar excels in hierarchical manipulation, while visualizations demonstrate knowledge retention and task discovery.
Project Page: \url{https://stellarvla.github.io/}.

\end{abstract}

\section{Introduction}
\label{sec:intro}
A core requirement for lifelong agents is continual learning that preserves prior knowledge while adapting to new situations. Prior work on robotic continual imitation learning (CIL) addresses catastrophic forgetting through skill expansion~\cite{wan2024lotus} or parameter-efficient adaptation~\cite{TAIL,IsCiL}, demonstrating promising performance in small- to medium-scale policy models. Recent advances in VLA models~\cite{UniVLA, pi0,pi05} demonstrate strong task learning capabilities in accumulating task experience from large-scale robotic data. However, adapting conventional CIL methods to VLAs introduces significant training and storage overhead, due to their reliance on increasing adapters or task-specific modules.

Encouragingly, recent work~\cite{liu2026pretrained} shows that VLA models can achieve strong continual learning performance with simple experience replay~\cite{ER} strategy. While promising, this result also challenges the necessity of increasingly complex CIL algorithmic designs. At this intersection between VLA scaling and conventional CIL methods, we seek to further investigate whether VLA models can reduce both the dependence on large-scale pretraining data and data replay during continual learning, thereby lowering training cost and storage overhead toward more scalable lifelong learning.

We argue that effective continual learning in VLA models should focus on enabling the model to capture task and skill relationships from data. Recent task-centric representation learning methods~\cite{VQBET, mete2024quest, VQ-VLA} have demonstrated improved policy adaptation, while LEGION~\cite{legion} extends this paradigm to state-based continual reinforcement learning for incremental skill acquisition.
Meanwhile, Mixture-of-Experts (MoE) architectures~\cite{shazeer2017outrageously} enable dynamic routing across specialized experts, offering a mechanism to mitigate task interference. These insights motivate a key question: \textit{Can VLA models learn latent task knowledge for continual policy evolution via adaptive specialization?}

To explore this question, we propose Stellar VLA, an end-to-end framework for CIL in vision-language-based manipulation. 
Building upon Dirichlet-process-based knowledge modeling and alternating optimization~\cite{legion}, we develop two variants for continual task and skill acquisition. T-Stellar employs a Dirichlet Process Mixture Model (DPMM) for task-centric modeling, while TS-Stellar extends it with a Hierarchical Dirichlet Process (HDP) to capture reusable skills across tasks. Alternating updates between task representations and knowledge distributions enable their co-evolution. We further integrate the learned knowledge into the policy through task-knowledge-guided expert routing, conditioned on knowledge relation embeddings and Top-$K$ semantic embeddings. This design balances parameter sharing and specialization without increasing model size.

Stellar VLAs are evaluated in both simulation and real-world settings. On the LIBERO benchmark, T-Stellar and TS-Stellar outperform VLA~\cite{MoDE, UniVLA, pi0, pi05} and CIL~\cite{ER, hu2022lora, wan2024lotus, IsCiL} baselines under both scratch and pretrained settings with only 1\% data replay, demonstrating strong continual learning performance. 
Real-world dual-arm experiments further show effective cross-embodiment adaptation, with TS-Stellar excelling in modeling complex hierarchical actions. Visualizations demonstrate skill acquisition while retaining prior knowledge. To conclude our contributions:

\begin{itemize}
    \item We present \textbf{Stellar VLA}, a CIL framework that models a Dirichlet Process–based knowledge space and jointly optimizes task representations and knowledge distribution via a self-evolving learning scheme. Two variants are introduced: \textbf{T-Stellar} using DPMM for Task-centric modeling, and \textbf{TS-Stellar} extending to HDP for hierarchical Task–Skill structure.

    \item We propose knowledge-posterior-guided expert routing using knowledge relation and Top-$K$ semantic embeddings, enabling efficient specialization without increasing model size.

    \item Experiments on LIBERO and real-world tasks demonstrate robust improvements in task knowledge adaptation, with consistent gains across different embodiments, while both variants capture evolving task or skill structure.
\end{itemize}

\begin{figure}[t]
    \centering
    \includegraphics[width=1\linewidth]{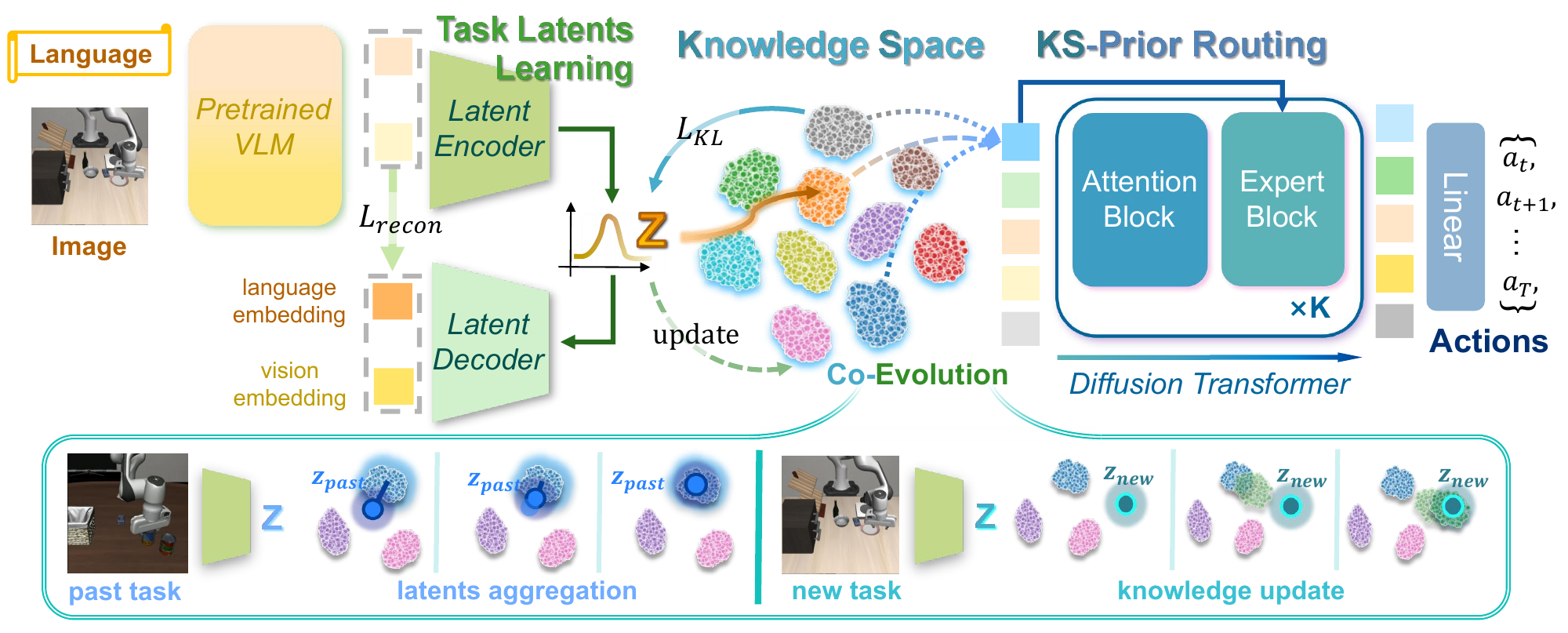}
    \caption{\textbf{Overall architecture of Stellar VLA.} CLIP~\cite{radford2021learning} and FiLM~\cite{perez2018film}-conditioned ResNet encode language and visual inputs respectively. Task-centric representation $z$ and knowledge space are jointly learned through knowledge update and latents aggregation, as detailed in Section~\ref{sec:k-learn}. The learned knowledge prior guides the MoE action head for motion prediction, as detailed in Section~\ref{sec:moe}.}
    \label{fig:arch}
\end{figure}
\section{Related Work}
\label{sec:related_work}

\noindent\textbf{Vision-Language-Action Models.}
Vision-language-action (VLA) models learn unified policies by jointly modeling perception, language, and control from robotic data. Recent works such as RT-2~\cite{zitkovich2023rt} and OpenVLA~\cite{kim2024openvla} demonstrate strong cross-task generalization, and more recent advances including $\pi_0$ and $\pi_{0.5}$~\cite{pi0,pi05}, further improve scalability and efficiency, where $\pi_{0.5}$ additionally leverages auxiliary multimodal data for representation learning. These advances make VLA a promising paradigm for general-purpose robot learning, but also raise challenges for continual adaptation.

\noindent\textbf{Continual Learning in Robotics.}
To address catastrophic forgetting in robotic policy learning, continual imitation learning (CIL) has been widely studied. Early methods are typically based on experience replay~\cite{ER} and parameter isolation~\cite{mallya2018packnet}. Recent work focuses on structured policy learning, including hierarchical skill expansion~\cite{wan2024lotus,lee2025policy,zhang2026atomicvla} and parameter-efficient adaptation~\cite{TAIL,IsCiL,SDP,lei2025dynamic,romer2026clare}.
However, these methods typically introduce additional data processing requirements or model parameters, leading to higher data and training costs when scaling to VLA models, which often include over 1B parameters and require large-scale training.

Recent work~\cite{yu2026lifelong,liu2026pretrained} shows that pretrained VLA models~\cite{pi0} can achieve strong CIL performance with simple replay using only 20\% data or latent space rehearsal. This suggests promising scalability, yet reducing storage while improving task knowledge adaptation remains an open challenge. We address this by proposing a fixed-size VLA model that enables efficient CIL with only 1\% data replay.

\noindent\textbf{Task-centric Representation Learning.}
Learning task-centric representations has become a promising direction for improving cross-task learning. Existing approaches, mainly studied in multi-task settings, model task structure through compact action representations~\cite{jiang2022efficient,VQBET,VQ-VLA} or vision-language abstractions~\cite{wang2024omnijarvis,wu2023unleashing,zhang2025dreamvla,hu2023thought,belkhale2024rt, deng2025best3dscenerepresentation}, with UniVLA~\cite{UniVLA} exemplifying the latter.

Recent work further explores more flexible structures for task adaptation, including grounded task expansion~\cite{VLA-OS} and action-centric incremental modeling~\cite{zheng2025imanip,yao2025think}, but often relies on manual annotations or parameter expansion. LEGION~\cite{legion} introduces a non-parametric knowledge space for state-based lifelong reinforcement learning. Inspired by LEGION, we adopt a self-evolving task knowledge space for vision-language continual imitation learning and extend it with hierarchical task--skill modeling and knowledge-guided expert routing.

\section{Method}
In this section, we present the architecture of Stellar VLA (Figure~\ref{fig:arch}), which enables CIL for vision-language manipulation. Our goal is to continually evolve task-relevant knowledge, enabling Stellar VLA to acquire new skills while mitigating forgetting of prior tasks. We first define the CIL setting, then describe the design and evolution of the knowledge space, and finally explain the process of guiding the expert policy via knowledge priors.

\subsection{Problem Formulation} \label{sec: PF}
In the CIL setting, an agent learns sequentially from a stream of tasks $\{\mathcal{T}_j\}_{j=1}^\infty$, where each task $j$ provides $N$ expert demonstrations $\{\tau_j^n\}_{n=1}^N$, consisting of language instructions $\boldsymbol{l}$, observations $\boldsymbol{o}$, and actions $\boldsymbol{a}$.
Unlike multi-task learning assuming full access to all demonstrations, CIL requires acquiring new task knowledge under limited access to past data while retaining previously learned skills, reflecting the realistic demands of lifelong robot learning.

To alleviate forgetting caused by limited access to past tasks, we adopt Experience Replay (ER) with a small buffer $\mathcal{B}$. After completing task $j$, $q\%$ of its demonstrations are stored in $\mathcal{B}$ and jointly trained with current data when learning task $j{+}1$.
However, relying on only a small portion of past data often leads to prediction drift on previous tasks $\mathcal{T}_k$ ($k < j{+}1$). To address this, we design a VLA architecture that continually discovers and preserves task knowledge, enabling dynamic expert composition for better parameter sharing and reduced interference.

\subsection{Dirichlet-Process-based Knowledge Space}\label{sec:dp}

To capture structured and interpretable task knowledge, we go beyond task latents and organize them into a high-level, dynamically evolving knowledge space. While latent features encode task-relevant variations, they lack explicit task identity, making it difficult to distinguish tasks during continual learning. To model such an evolving space, we adopt a Dirichlet Process (DP) prior, defined as $G \sim DP(\alpha, G_0)$, which supports clustering with an unbounded number of components. Here, $G_0$ denotes the base distribution over latent knowledge components, and $G$ defines a stochastic measure over task or skill clusters. Samples from $G$ induce clustering via shared components while allowing new components to emerge as tasks evolve. Building on Dirichlet Process, we design two knowledge spaces: a task-centric space and a hierarchical task-skill space.

\noindent\textbf{Task-Centric Knowledge Modeling from DPMM.}
Following LEGION's task-level knowledge modeling~\cite{legion}, we instantiate a Dirichlet Process Mixture Model (DPMM) to model task-level structure in the latent space. Based on Dirichlet-Process sampling, task latent $z_j$ in task $j$ under the task-set probability measure $G$ can be expressed as $ z_j \sim F_{task}(\theta_j), \theta_j \sim G$, where $\theta_j$ denotes task-specific distribution parameters and $F_{task}$ is the observation model. Assuming Gaussian components, this formulation induces a dynamic clustering of task representations in the knowledge space.

\noindent\textbf{Hierarchical Task-Skill Knowledge Modeling from HDP.}
The DPMM-based formulation models tasks via cluster assignments under a shared global mixture, but robotic tasks often share reusable subskills. To address this, we adopt a Hierarchical Dirichlet Process (HDP)~\cite{teh2006hierarchical} to model shared skill components across tasks. Specifically, each task $j$ is modeled as a distribution over skills indexed by $i$, where skill-level latent variables $z_{ji}$ are generated via:
\begin{equation}
\label{eq:hdp2}
z_{ji} \sim F_{skill}(\theta_{ji}), \quad \theta_{ji} \sim G_j, \quad G_j \sim DP(\gamma, G), \quad G \sim DP(\alpha, G_0)
\end{equation}

Here, $z_{ji}$ denotes the latent representation of skill $i$ in task $j$, $\theta_{ji}$ is its corresponding distribution parameter, and $G_j$ is the task-specific measure drawn from skill-set measure $G$. Under diagonal Gaussian assumptions, each skill is parameterized as $\theta_i^{skill} = (\mu_i^{skill}, \sigma_i^{skill})$. Given a task $j$, its overall distribution parameter $\theta_j^{task} = (\mu_j^{task}, \sigma_j^{task})$ is obtained by aggregating its skill components:
\begin{equation}
\label{eq:hdptask}
\boldsymbol{\mu}_j^{\text{task}} 
= \sum_{i \in \text{task } j} \pi_{ji}\,\boldsymbol{\mu}_i^{\text{skill}}, 
\quad
\boldsymbol{\sigma}_j^{\text{task}} 
= \sqrt{
\sum_{i \in \text{task } j} \pi_{ji}\,
\Big[
(\boldsymbol{\sigma}_i^{\text{skill}})^{\odot 2}
+ (\boldsymbol{\mu}_i^{\text{skill}} - \boldsymbol{\mu}_j^{\text{task}})^{\odot 2}
\Big]
}
\end{equation}

Here, $\pi_{ji}$ is the mixture weight of skill $i$ in task $j$, $\odot 2$ denotes element-wise squaring. The task-level measure $G_j$ shares atoms with the global measure $G$ but uses task-specific weights, enabling skill sharing across tasks and forming a hierarchical task–skill knowledge space.

\subsection{Knowledge Space Learning via Self-Evolution}\label{sec:k-learn}
\begin{algorithm}[t]
\caption{Joint Evolution of Task-centric Latents and Knowledge Space}
\label{pse}
\begin{algorithmic}[1]
    \State Initialize knowledge-space distribution $\Theta$ using Dirichlet-Process model
    \State Initialize VAE encoder and decoder parameters $\phi, \psi$
    \State Initialize knowledge buffer $\mathcal{B}_{know}$
   
    \For{data $(\boldsymbol{l}_j,\boldsymbol{o}_j,\boldsymbol{a}_j)$ at iteration $t$}
        \State $z_j \sim q_\phi(z_j \mid \boldsymbol{l}_j,\boldsymbol{o}_j)$ 
        \State Compute $L_{\text{recon}}(q_\psi(z_j), \boldsymbol{l}_j,\boldsymbol{o}_j), L_{\text{KL}}(z_j,\Theta)$
        \State Update VAE parameters  \quad 
        $ \phi, \psi \leftarrow \phi, \psi - \eta \nabla_{\phi,\psi}\big(L_{\text{recon}} + L_{\text{kl}} \big)$
        
        \State Update knowledge buffer \quad $\mathcal{B}_{know} \gets \mathcal{B}_{know} \cup \{(\boldsymbol{l}_j,\boldsymbol{o}_j)\}$
        \If{$t \bmod N_{\text{dp}} = 0$ and $t < N_{\text{max}}$}
            \State Sample $K_{\text{know}}$ latent points $\{z_i\}_{i=1}^{K_{\text{know}}} \sim q_\phi(z \mid (\boldsymbol{l}_j,\boldsymbol{o}_j)),\; (\boldsymbol{l}_j,\boldsymbol{o}_j) \sim \mathcal{B}_{know}$
            \State Update knowledge parameters $\Theta$ using $\{z_i\}_{i=1}^{K_{\text{know}}}$
        \EndIf
    \EndFor
     
\end{algorithmic}
\end{algorithm}
We present a unified learning framework where task-centric representations and Dirichlet-Process-based knowledge space co-evolve during continual imitation learning, as summarized in Algorithm~\ref{pse}, inspired by LEGION~\cite{legion}. Following Section~\ref{sec:dp}, we consider the dependency chain from observations $(\boldsymbol{l},\boldsymbol{o})$ through latent representations $z$ to the knowledge space $\Theta$. To finally get stable knowledge features, a hierarchical variational inference is performed: task-centric latent representations are inferred via VAE~\cite{kingma2013auto} for new task learning, and task knowledge is inferred from them through Dirichlet-Process-based models to manage old and new task information. This process gives rise to two variants, T-Stellar and TS-Stellar, corresponding to task-centric and task-skill hierarchical knowledge spaces.

\noindent\textbf{Task-Centric Representation Learning.}
We learn task-centric representations $z_j$ from vision-language abstractions using a VAE trained on vision-language data (lines 5-7 in Algorithm~\ref{pse}). The reconstruction loss $L_{\text{recon}}$ ensures semantic consistency, while a Dirichlet-Process-aware KL term regularizes the latent space to align with the current knowledge clusters.

For T-Stellar, the KL term is computed as
$L_{\text{KL}_j} = \sum_{k=1}^{K_c} p_{jk} L_{\text{KL}_{jk}}$,
where $p_{jk}$ denotes the probability of assigning $z_j$ to cluster $k$, and $K_c$ is the number of task knowledge clusters.

For TS-Stellar, we extend the representation to a hierarchical VAE where task- and skill-level latents are jointly modeled. The reconstruction loss becomes:
\begin{equation}
L_{\text{recon}}^\text{TS}
= L_{\text{recon}}(z_j^{task}, \boldsymbol{l}_j)
+ L_{\text{recon}}(z_j^{skill}, \boldsymbol{o}_j),
\end{equation}
where task latents $z_j^{task}$ decode language goals $\boldsymbol{l}$ and skill latents $z_j^{skill}$ decode visual observations $\boldsymbol{o}$.
The corresponding KL regularization follows the HDP structure in Eq.~\ref{eq:hdptask}:

\begin{equation}
L_{\text{KL}_j}^{\text{TS}}
= L_{\text{KL}_j}(z_j^{task}, \theta^{task})
+ L_{\text{KL}_j}(z_j^{skill}, \theta^{skill})
\end{equation}

\noindent\textbf{Knowledge Space Distribution Learning.}
To obtain knowledge distribution $\Theta$ (lines 8-11 in Algorithm~\ref{pse}), latent representations $\{z_j\}$ are treated as observations of $G$, and variational inference is performed to approximate the posterior $p(\Theta, G \mid \{z_j\})$. 
Memoized variational Bayes (memoVB)~\cite{hughes2013memoized} is adopted for efficient incremental updates, enabling global statistic sharing across tasks.

The task-centric representation learning provides latent features to update the knowledge distribution, while the knowledge space, via KL loss $L_{kl}$, constrains $z_j$ to infer under task or skill hypotheses. This design establishes a self-evolving cycle that autonomously retains and discovers task knowledge.

\subsection{Action-Knowledge-guided Expert Routing }
\label{sec:moe}
\begin{figure}[t]
    \centering
    \includegraphics[width=0.95\linewidth]{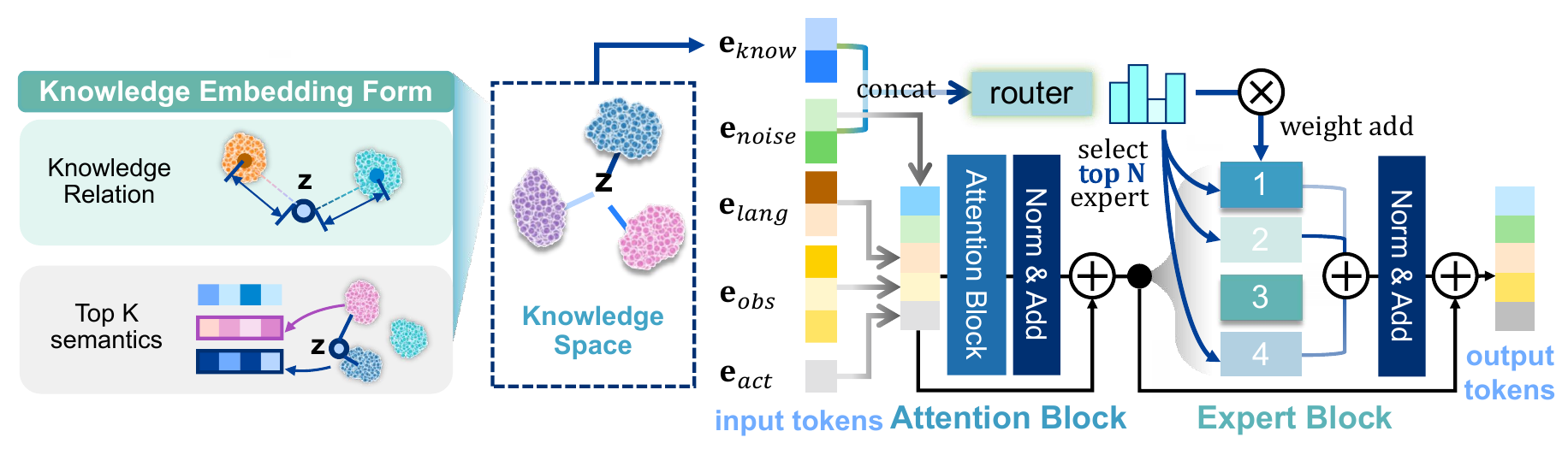}
    \caption{\textbf{Knowledge-prior-routed MoE action head.} Two knowledge embeddings, relation and top-K semantic, are computed for expert routing, alongside language, noise, observation and noise action tokens fed into the denoising transformer.}
    \label{fig:MoE}
\end{figure}

After learning the knowledge space and task-centric latents, we introduce a knowledge-routed MoE module to allocate task-specific action prediction parameters, minimizing interference between unrelated tasks while enhancing sharing among related ones.
A diffusion-based MoE action head is adopted referring to MoDE~\cite{MoDE}:
\begin{equation}
\label{eq:moe-layer}
\text{MoE}(\mathbf{e}, \mathbf{f}_{ro}) = \sum_{i=1}^{N_e} \mathbf{Router}_i(\mathbf{f}_{ro}) \cdot \mathbf{Expert}_i(\mathbf{e})
\end{equation}
Here, $\mathbf{e}$ is the input token feature tensor and $\mathbf{f}_{ro}$ the routing features, defined in MoDE as the noise embedding $\mathbf{f}_{ro}=\mathbf{e}_{noise} = \phi(\sigma_{\mathbf{t}})$, where $\sigma_{\mathbf{t}}$ is the denoising level. Unlike MoDE’s noise-level routing, we guide expert allocation via the knowledge space, enabling continual task learning with task-specific parameter sharing and differentiation.
Since DPMM and HDP produce variable numbers of clusters, directly feeding their task distributions into a fixed-dimension network is impractical. Thus, we design two knowledge embedding formats, as shown in Figure~\ref{fig:MoE}.

\textbf{Knowledge Relation Embedding} measures the distance between latent $z$ and cluster centers $\mu_k$. The membership probability $p_k$, obtained from the knowledge distribution, denotes the posterior assignment of $z$ to cluster $k$, and is used to compute the relation embedding as:
\begin{equation}
\mathbf{f}_{\text{R}} = \sum_{k=1}^{K_c} p_k \, |z - \mu_k|
\end{equation}
\textbf{Top-K Semantic Embedding} represents the discrete skill semantic affiliation of $z$ by aggregating learnable embeddings of the top-K assigned clusters:
\begin{equation}
\mathbf{f}_{\text{S}} = \sum_{k \in \mathcal{I}_T} p_k \cdot \text{Embed}(k),
\mathcal{I}_T = \text{TopKIndice}({p_1, \dots, p_K}, z)
\end{equation}
Finally, the embedded knowledge of current task and skill could be represented as:
\begin{equation}
\mathbf{f}_{\text{know}} = 
\left[\,z\,  \| \mathbf{f}_{\text{R}} \; \| \; \mathbf{f}_{\text{S}} \,\right]
\end{equation}
The knowledge embedding $\mathbf{f}_{\text{know}}$ is introduced as an additional guidance token $\mathbf{e}_{know}$, which, together with $\mathbf{e}_{noise}$, are concatenated as input to the router for task-aware expert allocation.

By incorporating knowledge space guidance into expert routing, the model could achieve effective CIL through balanced parameter sharing and isolation, enabling related tasks to benefit from shared experience while preventing catastrophic forgetting across dissimilar ones.

\section{Experiments}
Our experiments address four questions: 
(I) How do T-Stellar and TS-Stellar compare with existing VLA and CIL methods? 
(II) What advantages does the hierarchical task-skill knowledge space in TS-Stellar provide over task-centric modeling in T-Stellar? 
(III) How do task/skill modeling and knowledge-guided expert routing jointly contribute to continual learning performance?
(IV) How does the task knowledge space evolve to discover and preserve skills over time?
\subsection{Simulation Evaluation } \label{sec:main_sim}
We evaluate our models on the LIBERO benchmark~\cite{liu2023libero}, including LIBERO-goal, LIBERO-long, and LIBERO-30*, a 30-task subset of LIBERO-90, covering diverse goals, long-horizon reasoning, and many tasks settings. LIBERO is designed for single-arm manipulation, and demonstrations include static and wrist camera images, with actions defined as end-effector poses and gripper states. We compare against both advanced VLA models and robotic CIL methods.
\begin{table*}[t]
\caption{\textbf{Overall performance on the LIBERO benchmark.} Baselines are grouped into pretrained and scratch categories. Results are reported in percentages. The highest performance is highlighted in \textbf{bold} and the second highest is \underline{underline}.} 
\centering
\small
\setlength{\tabcolsep}{1.5pt}
\renewcommand{\arraystretch}{1}
\resizebox{1\columnwidth}{!}{
\begin{tabular}{ll|ccc|cc|cccc|cc}
\toprule
\multirow{2}{*}{\textbf{tasks}} & \multirow{2}{*}{\textbf{metrics}} 
&\multicolumn{5}{c|}{Scratch VLA Models} &\multicolumn{6}{c}{Pretrained VLA Models (on cross-embodiment data)} \\
&& MoDE~\cite{MoDE}&UniVLA~\cite{UniVLA} &$\pi_0$~\cite{pi0} &
\textbf{T-Stellar} & \textbf{TS-Stellar} & MoDE~\cite{MoDE}&UniVLA~\cite{UniVLA} &$\pi_0$~\cite{pi0} &$\pi_{0.5}$~\cite{pi05} &
\textbf{T-Stellar} & \textbf{TS-Stellar}\\ 
\midrule

\multirow{4}{*}{\textbf{LIBERO-goal}} 
& FWT (↑) & 71.0\std{2.1} & 3.1\std{0.7} & \textbf{82.9\std{3.9}} & \cellcolor{gray!20}\underline{81.4\std{0.2}} & \cellcolor{gray!20}80.2\std{2.3} &69.7\std{1.8} &\textbf{87.5\std{0.1}} &74.9\std{2.7} &\underline{84.5\std{2.5}} &\cellcolor{gray!20}76.6\std{3.4}&\cellcolor{gray!20}73.6\std{4.1}\\ 

& NBT (↓) &36.2\std{2.8} &\textbf{3.1\std{0.7}} &49.0\std{1.0} &\cellcolor{gray!20}\underline{20.7\std{2.8}} &\cellcolor{gray!20}22.2\std{3.3} &\underline{23.2\std{2.9}} &65.7\std{6.1} &25.6\std{1.0} &38.4\std{3.7} &\cellcolor{gray!20}\textbf{19.1\std{3.6}} &\cellcolor{gray!20}28.0\std{3.7}\\ 

& AUC (↑)  &39.7\std{1.0} &0.4\std{0.2} &45.5\std{3.1}  &\cellcolor{gray!20}\textbf{61.7\std{2.2}} &\cellcolor{gray!20}\underline{60.7\std{1.2}} &46.8\std{2.9} &35.2\std{4.7} &\underline{55.6\std{3.3}} &55.4\std{4.9} & \cellcolor{gray!20}\textbf{57.5 \std{1.1}} &\cellcolor{gray!20}48.1\std{2.2}\\ 

& Final SR (↑) &33.1\std{2.4} &0.0\std{0.0} &35.7\std{0.5} &\cellcolor{gray!20}\textbf{67.9\std{0.5}} &\cellcolor{gray!20}\underline{64.2\std{1.5}} &41.6\std{4.8} &20.5\std{4.7} &41.1\std{1.1} &43.8\std{3.8} &\cellcolor{gray!20}\textbf{57.0\std{2.0}}&\cellcolor{gray!20}\underline{44.1\std{5.2}} \\ 
\midrule

\multirow{4}{*}{\textbf{LIBERO-long}} 

& FWT (↑) & 71.6\std{3.6} & 6.1\std{0.1} & 57.3\std{3.1} & \cellcolor{gray!20}\textbf{76.3\std{4.0}} & \cellcolor{gray!20}\underline{75.5\std{0.6}} & 71.1\std{1.5} & 70.3\std{5.6} & 60.5\std{2.6} & 55.5\std{3.3} & \cellcolor{gray!20}\underline{80.3\std{1.6}} & \cellcolor{gray!20}\textbf{81.9\std{4.1}} \\

& NBT (↓) & 46.3\std{1.6} & \textbf{6.0\std{0.1}} & 48.7\std{3.7} & \cellcolor{gray!20}41.6\std{4.1} & \cellcolor{gray!20}\underline{37.3\std{2.1}} & 46.4\std{1.1} & \textbf{31.3\std{4.0}} & \underline{34.4\std{5.5}} & 34.8\std{1.9} & \cellcolor{gray!20}38.0\std{2.5} & \cellcolor{gray!20}37.6\std{4.4} \\

& AUC (↑) & 34.3\std{4.2} & 0.9\std{0.0} & 22.2\std{1.0} & \cellcolor{gray!20}\underline{41.2\std{0.7}} & \cellcolor{gray!20}\textbf{43.6\std{1.9}} & 32.9\std{1.4} & 43.6\std{2.4} & 35.6\std{3.7} & 29.8\std{5.9} & \cellcolor{gray!20}\underline{48.6\std{2.6}} & \cellcolor{gray!20}\textbf{50.8\std{1.5}} \\

& Final SR (↑) & 20.2\std{2.9} & 0.0\std{0.0} & 12.0\std{2.2} & \cellcolor{gray!20}\underline{34.2\std{2.1}} & \cellcolor{gray!20}\textbf{35.0\std{3.2}} & 19.9\std{1.4} & 17.1\std{2.0} & 24.3\std{5.1} & 21.6\std{5.6} & \cellcolor{gray!20}\underline{32.2\std{4.4}} & \cellcolor{gray!20}\textbf{41.7\std{3.0}} \\

\midrule

\multirow{4}{*}{\textbf{LIBERO-30 *}} 
& FWT (↑) & 48.2 & 35.5 & 72.7 & \cellcolor{gray!20}\textbf{79.6} & \cellcolor{gray!20}\underline{73.3} & 52.9 & \textbf{87.9} & 70.3 & 75.0 & \cellcolor{gray!20}43.8 & \cellcolor{gray!20}\underline{84.8} \\

& NBT (↓) & \textbf{15.4} & 28.1 & 46.9 & \cellcolor{gray!20}31.0 & \cellcolor{gray!20}\underline{27.8} & \underline{11.9} & 57.7 & 33.5 & 41.7 & \cellcolor{gray!20}\textbf{4.1} & \cellcolor{gray!20}37.2 \\

& AUC (↑) & 33.6 & 9.4 & 31.1 & \cellcolor{gray!20}\textbf{49.7} & \cellcolor{gray!20}\underline{48.3} & \underline{43.0} & 36.5 & 40.8 & 37.9 & \cellcolor{gray!20}38.3 & \cellcolor{gray!20}\textbf{50.8} \\

& Final SR (↑) & 28.5 & 12.0 & 24.9 & \cellcolor{gray!20}\textbf{42.9} & \cellcolor{gray!20}\underline{42.6} & \underline{40.3} & 29.6 & 34.5 & 35.3 & \cellcolor{gray!20}35.8 & \cellcolor{gray!20}\textbf{41.0}\\
\bottomrule
\end{tabular}
}
\vspace{1ex} 

\setlength{\tabcolsep}{1.5pt}
\renewcommand{\arraystretch}{1}
\resizebox{1\columnwidth}{!}{
\begin{tabular}{ll|cccc|cc|cccc|cc}
\toprule
\multirow{2}{*}{\textbf{tasks}} & \multirow{2}{*}{\textbf{metrics}} 
&\multicolumn{6}{c|}{Scratch CIL Methods} &\multicolumn{6}{c}{Pretrained CIL Methods (on LIBERO-90)} \\
&& ER~\cite{ER}&SeqLoRA &LoTUS~\cite{wan2024lotus} &IsCiL~\cite{IsCiL}&
\textbf{T-Stellar} & \textbf{TS-Stellar} & ER~\cite{ER}&SeqLoRA &LoTUS~\cite{wan2024lotus} &IsCiL~\cite{IsCiL}&
\textbf{T-Stellar} & \textbf{TS-Stellar}\\ 
\midrule

\multirow{4}{*}{\textbf{LIBERO-goal}} 

& FWT (↑) 

& 70.8\std{1.5} & 26.2\std{4.6} & 44.8\std{0.3} & 24.3\std{5.8} 

& \cellcolor{gray!20}\textbf{81.4\std{0.2}} & \cellcolor{gray!20}\underline{80.2\std{2.3}} 

& 80.3\std{2.3} & 80.3\std{0.6} & 46.8\std{4.3} & 66.8\std{1.3} 

& \cellcolor{gray!20}\underline{80.4\std{0.3}} & \cellcolor{gray!20}\textbf{82.7\std{1.6}} \\ 

& NBT (↓) 

& 33.0\std{2.0} & 23.9\std{4.0} & 31.3\std{6.3} & \textbf{3.0\std{1.2}} 

& \cellcolor{gray!20}\underline{20.7\std{2.8}} & \cellcolor{gray!20}22.2\std{3.3} 

& 21.1\std{0.6} & 78.8\std{1.5} & 37.7\std{3.9} & \textbf{7.8\std{2.5}} 

& \cellcolor{gray!20}\underline{12.6\std{1.0}} & \cellcolor{gray!20}19.4\std{2.2} \\ 

& AUC (↑) 

& 41.8\std{0.4} & 8.9\std{1.1} & 22.2\std{5.0} & 22.0\std{2.8} 

& \cellcolor{gray!20}\textbf{61.7\std{2.2}} & \cellcolor{gray!20}\underline{60.7\std{1.2}} 

& 60.5\std{1.8} & 26.2\std{1.0} & 18.5\std{1.5} & 60.9\std{2.8} 

& \cellcolor{gray!20}\textbf{66.7\std{0.5}} & \cellcolor{gray!20}\underline{64.8\std{1.6}} \\ 

& Final SR (↑) 

& 34.2\std{0.9} & 2.4\std{0.8} & 16.3\std{4.8} & 18.5\std{0.5} 

& \cellcolor{gray!20}\textbf{67.9\std{0.5}} & \cellcolor{gray!20}\underline{64.2\std{1.5}} 

& 55.3\std{1.0} & 9.5\std{0.1} & 10.8\std{1.1} & 48.5\std{2.0} 

& \cellcolor{gray!20}\textbf{62.1\std{1.5}} & \cellcolor{gray!20}\underline{57.3\std{5.1}} \\ 

\midrule

\multirow{4}{*}{\textbf{LIBERO-long}}

& FWT (↑) 

& 70.8\std{6.2} & 9.6\std{6.1} & 11.5\std{3.0} & 26.0\std{6.5} 

& \cellcolor{gray!20}\textbf{76.3\std{4.0}} & \cellcolor{gray!20}\underline{75.5\std{0.6}} 

& \underline{80.3\std{0.3}} & 62.0\std{1.3} & 32.3\std{2.3} & 65.8\std{0.3} 

& \cellcolor{gray!20}79.3\std{2.1} & \cellcolor{gray!20}\textbf{81.5\std{1.3}} \\ 

& NBT (↓) 

& 47.3\std{5.5} & 8.8\std{5.4} & \textbf{0.8\std{1.7}} & \underline{15.0\std{4.7}} 

& \cellcolor{gray!20}41.6\std{4.1} & \cellcolor{gray!20}37.3\std{2.1} 

& 48.4\std{2.0} & 56.0\std{1.5} & \textbf{2.8\std{2.7}} & \underline{31.5\std{0.3}} 

& \cellcolor{gray!20}36.8\std{2.8} & \cellcolor{gray!20}37.7\std{2.0} \\ 

& AUC (↑) 

& 31.0\std{3.0} & 2.3\std{1.4} & 11.3\std{4.2} & 15.2\std{3.8} 

& \cellcolor{gray!20}\underline{41.2\std{0.7}} & \cellcolor{gray!20}\textbf{43.6\std{1.9}} 

& 40.9\std{1.1} & 16.6\std{0.5} & 29.4\std{0.5} & 41.4\std{0.1} 

& \cellcolor{gray!20}\underline{48.3\std{3.9}} & \cellcolor{gray!20}\textbf{50.5\std{1.3}} \\ 

& Final SR (↑) 

& 18.9\std{1.1} & 0.8\std{0.6} & 10.5\std{5.5} & 6.8\std{0.8} 

& \cellcolor{gray!20}\underline{34.2\std{2.1}} & \cellcolor{gray!20}\textbf{35.0\std{3.2}} 

& 16.1\std{0.7} & 6.0\std{0.5} & 31.8\std{2.3} & 17.3\std{0.3} 

& \cellcolor{gray!20}\underline{36.3\std{3.8}} & \cellcolor{gray!20}\textbf{40.9\std{2.3}} \\ 
\bottomrule
\end{tabular}
}
\label{tab:LIBERO_results}
\end{table*}

\noindent\textbf{Baselines. } 
For VLA baselines, we select strong multitask performers on LIBERO and retrain them under the CIL setting for fair comparison. The baselines include: 1) MoDE~\cite{MoDE}, a diffusion-based Mixture-of-Experts model; 2) UniVLA~\cite{UniVLA}, which learns task-centric representations via latent action pretraining; and 3) $\pi_0$~\cite{pi0} and $\pi_{0.5}$~\cite{pi05}, which leverage large-scale robotic and heterogeneous data for generalization.
All methods are evaluated under both from-scratch and  cross-embodiment-data pretrained settings, except $\pi_{0.5}$, which is only evaluated in the pretrained setting, as its main contribution falls on web-scale data pretraining. All models are trained sequentially following the LIBERO task order with Experience Replay (1\% memory).
For CIL baselines, we include representative replay-, adapter-, and skill-expansion-based methods: 1) Experience Replay (ER)~\cite{ER}; 2) Sequential Low-Rank Adaptation (SeqLoRA)~\cite{hu2022lora}; 3) LoTUS~\cite{wan2024lotus}, with incremental skill expansion; and 4) IsCiL~\cite{IsCiL}, with retrieval-based adaptation. All methods are evaluated under both pretrained and scratch settings. Since LoTUS and IsCiL rely on in-domain pretraining, pretrained comparisons are conducted on LIBERO-90, and LIBERO-30* results are omitted.

\noindent\textbf{Metrics. }\label{sec:metric} 
Four metrics are used: FWT (forward transfer), NBT (negative backward transfer), AUC (area under the success rate curve)~\cite{IsCiL, wan2024lotus, diaz2018don, liu2023libero}, and Final SR (final success rate). FWT measures forward learning, NBT captures forgetting, AUC reflects overall stability, and Final SR denotes performance after training all tasks. Each policy is evaluated over 100 trials with 50 initial states. Results are reported as mean $\pm$ standard deviation over 3 training seeds for LIBERO-goal and -long. Due to the high cost of LIBERO-30*, only single-run results are reported.

\noindent\textbf{Results. }
Table \ref{tab:LIBERO_results} shows that T-Stellar and TS-Stellar achieve the best AUC and final success rates across all scenarios among both VLA and CIL baselines. Notably, in the scratch setting, our methods obtain over 50\% average improvement in AUC and Final SR relative to all baselines~\cite{UniVLA, MoDE, pi0, pi05, ER, wan2024lotus, IsCiL}. Some methods achieve lower NBT mainly due to very low FWT, reflecting the inherent trade-off between forward transfer and backward transfer in continual learning. Among pretrained VLAs, $\pi_0$~\cite{pi0} ($\sim$ 3B), $\pi_{0.5}$~\cite{pi05} ($\sim$ 3B), and UniVLA~\cite{UniVLA} ($\sim$7B) show competitive results, likely due to large-scale parameters enabling near-orthogonal gradient updates across tasks in simpler settings. However, without explicit task-specific modeling, their performance is unstable across tasks, while our method achieves the best results with only $\sim$1B parameters. Compared with CIL methods, without parameter growth as in LOTUS~\cite{wan2024lotus} or IsCiL~\cite{IsCiL} and with minimal replay data, we achieve 20\% improvement in Final SR over all baselines, demonstrating the effectiveness of self-supervised knowledge evolution with task-guided action allocation. 
Notably, pretraining on LIBERO-90 yields more stable performance than heterogeneous pretraining due to better alignment in embodiment and task distribution, enabling efficient knowledge transfer with minimal adaptation. In contrast, pretraining on over 1k cross-embodiment tasks requires more parameter updates to fit new scenarios. Nevertheless, Stellar VLAs still achieve state-of-the-art performance under heterogeneous pretraining, demonstrating strong knowledge transfer capability.

In response to Question (I), our methods outperform both from-scratch and pretrained baselines, demonstrating strong knowledge accumulation and transfer ability. TS-Stellar further excels on long-horizon tasks, motivating deeper analysis of its hierarchical modeling in the following experiments.

\subsection{Real-World Evaluation on Dual-Arm Manipulation}
\begin{table*}[t]
\caption{\textbf{Performance on real-world experiments.} The highest performance is highlighted in \textbf{bold} and the second highest is \underline{underline}. }
\centering
\small
\setlength{\tabcolsep}{5pt}
\renewcommand{\arraystretch}{1.2}
\resizebox{0.85\columnwidth}{!}{
\begin{tabular}{ll|c|cccc|cc}
\toprule
\textbf{tasks} & \textbf{metrics} &ER~\cite{ER}& MoDE~\cite{MoDE}&UniVLA~\cite{UniVLA} &$\pi_0$~\cite{pi0} &$\pi_{0.5}$~\cite{pi05} &
\textbf{T-Stellar} & \textbf{TS-Stellar} \\ 
\midrule

\multirow{4}{*}{\shortstack{\textbf{Dual arm}\\\textbf{real world}\\\textbf{task}}}

& FWT (↑) 

& \underline{97.1} & 15.7 & 70.0 & \textbf{98.6} & 95.7 

& \cellcolor{gray!20}\textbf{98.6} & \cellcolor{gray!20}\textbf{98.6} \\ 

& NBT (↓) 

& 21.9 & \textbf{3.2} & 37.1 & 34.6 & 25.8 

& \cellcolor{gray!20}12.4 & \cellcolor{gray!20}\underline{7.4} \\ 

& AUC (↑) 

& 79.9 & 13.0 & 43.6 & 72.6 & 75.6 

& \cellcolor{gray!20}\underline{89.6} & \cellcolor{gray!20}\textbf{93.4} \\ 

& Final SR (↑) 

& 70.0 & 10.0 & 21.4 & 57.1 & 72.9 

& \cellcolor{gray!20}\underline{84.3} & \cellcolor{gray!20}\textbf{90.0} \\
\bottomrule
\end{tabular}
}
\label{tab:rw}
\end{table*}

\begin{figure}[t]
    \centering
    \includegraphics[width=1\linewidth]{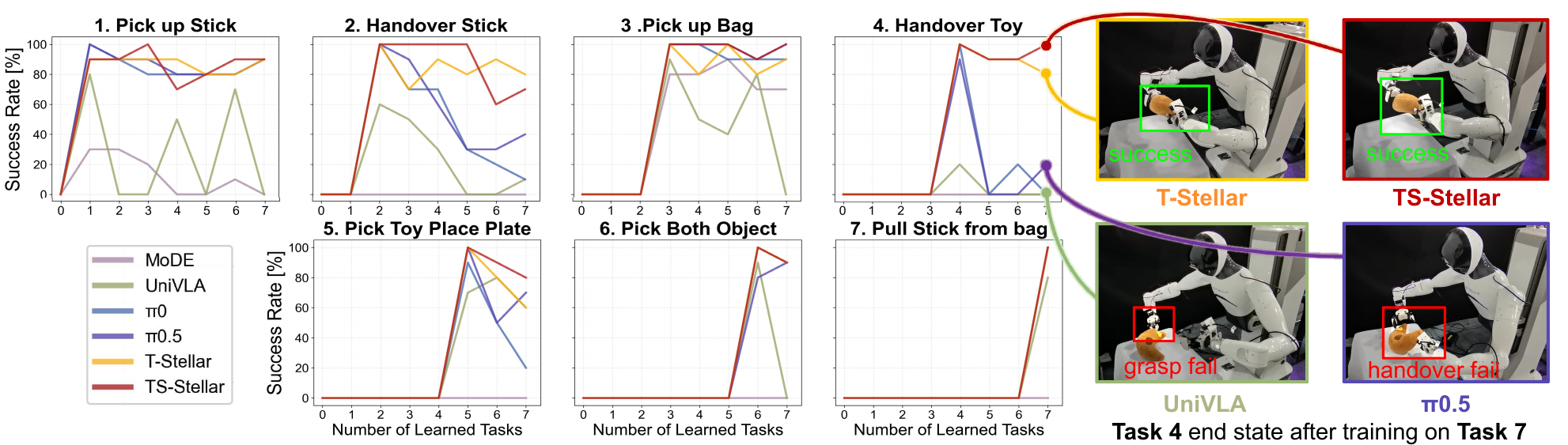}
    \caption{\textbf{Success rate curves of real-world experiments and behavior visualization on ``Handover Toy'' after training on ``Pull Stick from bag''.} Stellar VLAs maintain high overall success rates, while UniVLA~\cite{UniVLA} and the $\pi$ series~\cite{pi0,pi05} exhibit clear forgetting during continual learning. MoDE~\cite{MoDE}, lacking adaptation to dual-arm robots, fails to perform most tasks.}
    \label{fig:vis_beh}
\end{figure}

To further validate performance on a new embodiment with compositional manipulation, we conduct real-world dual-arm manipulation experiments on seven tasks involving both single-arm and bimanual operations: ``Pick up Stick'', ``Handover Stick'', ``Pick up Bag'', ``Handover Toy'', ``Pick Toy Place Plate'', ``Pick Both Object'', and ``Pull Stick from Bag''. Models are trained with a 5\% replay rate for stability and evaluated following Section~\ref{sec:metric}, with ten trials per task.

\noindent\textbf{Baselines. } 
We compare against all VLA methods~\cite{MoDE, UniVLA, pi0, pi05} reported in Section~\ref{sec:metric}, and include ER~\cite{ER} as a CIL baseline. As described in LoTUS~\cite{wan2024lotus} and IsCiL~\cite{IsCiL}, both methods rely on in-domain pretraining. However, our real-robot data involves dual-arm manipulation with substantially different embodiment and scene configurations from LIBERO. Therefore, they are not included in the comparison. This further highlights the advantage of VLA models in adapting to heterogeneous data.

\noindent\textbf{Results. } 
Table~\ref{tab:rw} show that Stellar VLAs outperform all baselines in AUC, and Final SR while achieve strong NBT and FWT, exhibiting superior CIL ability on dual-arm real-world tasks. TS-Stellar further surpasses T-Stellar across all metrics, suggesting that task-skill modeling is particularly effective for long-horizon skill sequences and complex action mappings, thus addressing Question (II). Figure ~\ref{fig:vis_beh} also visualize the success rate curves of VLA baselines, as well as the behavior of representative VLA models evaluated on “Handover Toy” after training on “Pull Stick from Bag”. Results show that our method more effectively mitigates catastrophic forgetting in dual-arm tasks. In “Handover Toy”, which requires bimanual skills, UniVLA~\cite{UniVLA} and $\pi_{0.5}$~\cite{pi05} exhibit failures at different subtask stages, whereas our method maintains accurate grasping and coordinated handover behaviors.

\begin{table*}[t]
\centering
\small
\setlength{\tabcolsep}{4pt}
\renewcommand{\arraystretch}{1.1}
\begin{minipage}[t]{0.42\textwidth}
\centering
\caption{\textbf{Replay budget analysis.}}
\par\smallskip
\resizebox{\linewidth}{!}{
\begin{tabular}{cl|ccc}
\toprule
\textbf{Buffer} & \textbf{metrics} & \textbf{ER} & \textbf{T-Stellar} & \textbf{TS-Stellar} \\
\midrule
\multirow{4}{*}{10\%} & FWT (↑) & 80.8\std{2.5} & 83.4\std{2.2} & \textbf{85.9\std{1.2}} \\
& NBT (↓) & 8.4\std{1.4} & \textbf{1.2\std{0.8}} & 2.4\std{0.2} \\
& AUC (↑) & 76.1\std{1.7} & \underline{82.3\std{1.8}} & \textbf{83.8\std{1.2}} \\
& Final SR (↑) & 70.4\std{1.1} & \underline{80.1\std{1.6}} & \textbf{80.2\std{1.4}} \\
\midrule
\multirow{4}{*}{5\%} & FWT (↑) & 74.6\std{4.2} & \underline{81.2\std{2.1}} & \textbf{84.8\std{0.9}} \\
& NBT (↓) & \textbf{3.2\std{1.7}} & \underline{5.5\std{1.3}} & 9.3\std{1.7} \\
& AUC (↑) & 70.4\std{1.5} & \underline{76.4\std{0.7}} & \textbf{77.3\std{1.9}} \\
& Final SR (↑) & 65.0\std{1.5} & \underline{72.1\std{1.8}} & \textbf{73.2\std{2.4}} \\
\midrule
\multirow{4}{*}{1\%} & FWT (↑) & 70.8\std{6.2} & \textbf{76.3\std{4.0}} & \underline{75.5\std{0.6}} \\
& NBT (↓) & 47.3\std{5.5} & \underline{41.6\std{4.1}} & \textbf{37.3\std{2.1}} \\
& AUC (↑) & 31.0\std{3.0} & \underline{41.2\std{0.7}} & \textbf{43.6\std{1.9}} \\
& Final SR (↑) & 18.9\std{1.1} & \underline{34.2\std{2.1}} & \textbf{35.0\std{3.2}} \\
\bottomrule
\end{tabular}\label{tab:replay_budget}
}
\end{minipage}
\hfill
\begin{minipage}[t]{0.57\textwidth}
\centering
\caption{\textbf{Knowledge space ablation.}}
\par\smallskip

\resizebox{\linewidth}{!}{
\begin{tabular}{ll|cc
|cc}
\toprule
\textbf{tasks} & \textbf{metrics} & 
\textbf{w/o VAE} & 
\textbf{w/o KS} &
\textbf{T-Stellar} & 
\textbf{TS-Stellar} \\ 
\midrule

\multirow{4}{*}{\textbf{LIBERO-goal}} 
& FWT (↑)      & 70.8\std{1.5} & 76.2\std{2.0} & \textbf{81.4\std{0.2}} & \underline{80.2\std{2.3}} \\
& NBT (↓)      & 33.0\std{2.0} & 23.2\std{0.2} & \textbf{20.7\std{2.8}} & \underline{22.2\std{3.3}} \\
& AUC (↑)      & 41.8\std{0.4} & 54.7\std{2.0} & \textbf{61.7\std{2.2}} & \underline{60.7\std{1.2}} \\
& Final SR (↑) & 34.2\std{0.9} & 49.8\std{4.3} & \textbf{67.9\std{0.5}} & \underline{64.2\std{1.5}} \\
\midrule

\multirow{4}{*}{\textbf{LIBERO-long}} 
& FWT (↑)      & 70.8\std{6.2} & 72.3\std{2.3} & \textbf{76.3\std{4.0}} & \underline{75.5\std{0.6}} \\
& NBT (↓)      & 47.3\std{5.5} & 45.3\std{1.8} & \underline{41.6\std{4.1}} & \textbf{37.3\std{2.1}} \\
& AUC (↑)      & 31.0\std{3.0} & 34.3\std{3.6} & \underline{41.2\std{0.7}} & \textbf{43.6\std{1.9}} \\
& Final SR (↑) & 18.9\std{1.1} & 21.4\std{1.3} & \underline{34.2\std{2.1}} & \textbf{35.0\std{3.2}} \\
\midrule

\multirow{4}{*}{\textbf{LIBERO-30*}} 
& FWT (↑) & 45.0 & 45.1 & \textbf{79.6} & \underline{73.3} \\
& NBT (↓) & \underline{12.0} & \textbf{4.9} & 31.0 & 27.8 \\
& AUC (↑) & 32.7 & 38.9 & \textbf{49.7} & \underline{48.3} \\
& SR (↑)  & 26.3 & 35.8 & \textbf{42.9} & \underline{42.6} \\

\bottomrule
\end{tabular}\label{tab:ablation_dpmoe}
}
\end{minipage}
\end{table*}

\begin{table*}[t]
\centering
\small
\setlength{\tabcolsep}{4pt}
\renewcommand{\arraystretch}{1.1}

\begin{minipage}[t]{0.54\textwidth}
\centering
\caption{\textbf{Comparison with LEGION-style model.} LEGION-style model shares the same backbone as Stellar, but only employs DPMM policy conditioning.}
\par\smallskip
\resizebox{\linewidth}{!}{
\begin{tabular}{ll|ccc}
\toprule
\textbf{tasks} & \textbf{metrics} & \textbf{LEGION-style} & \textbf{T-Stellar} & \textbf{TS-Stellar} \\
\midrule
\multirow{4}{*}{\textbf{LIBERO-goal}}
& FWT ($\uparrow$) & 75.2\std{2.0} & \textbf{81.4\std{0.2}} & \underline{80.2\std{2.3}} \\
& NBT ($\downarrow$) & \underline{21.3\std{3.2}} & \textbf{20.7\std{2.8}} & 22.2\std{3.3} \\
& AUC ($\uparrow$) & 54.6\std{1.5} & \textbf{61.7\std{2.2}} & \underline{60.7\std{1.2}} \\
& Final SR ($\uparrow$) & 54.4\std{2.0} & \textbf{67.9\std{0.5}} & \underline{64.2\std{1.5}} \\
\midrule
\multirow{4}{*}{\textbf{LIBERO-long}}
& FWT ($\uparrow$) & 75.4\std{4.0} & \textbf{76.3\std{4.0}} & \underline{75.5\std{0.6}} \\
& NBT ($\downarrow$) & 44.6\std{3.0} & \underline{41.6\std{4.1}} & \textbf{37.3\std{2.1}} \\
& AUC ($\uparrow$) & 37.9\std{0.3} & \underline{41.2\std{0.7}} & \textbf{43.6\std{1.9}} \\
& Final SR ($\uparrow$) & 23.0\std{2.1} & \underline{34.2\std{2.1}} & \textbf{35.0\std{3.2}} \\

\bottomrule
\end{tabular}\label{tab:legion_style}
}
\end{minipage}
\hfill
\begin{minipage}[t]{0.43\textwidth}
\centering
\caption{\textbf{Prior-guided routing ablation.} $\mathbf{f}_{\text{R}}$ is knowledge relation embedding, and $\mathbf{f}_{\text{S}}$ is top-K semantic embedding, as in Sec.~\ref{sec:moe}.} 
\par\smallskip
\resizebox{\linewidth}{!}{
\begin{tabular}{l|ccc|c}
\toprule
\textbf{T-Stellar} & w/o ($\mathbf{f}_{\text{R}} + \mathbf{f}_{\text{S}}$) & w/ $\mathbf{f}_{\text{R}}$ & w/ $\mathbf{f}_{\text{S}}$   & w/ ($\mathbf{f}_{\text{R}} + \mathbf{f}_{\text{S}}$) \\
\midrule
FWT (↑) & 75.4\std{4.0} & 73.5\std{1.8} & \textbf{78.3\std{2.3}} & 76.3\std{4.0} \\
NBT (↓) & 44.6\std{3.0} & \textbf{41.5\std{2.5}} & 49.9\std{1.3} & 41.6\std{4.1} \\
AUC (↑) & 37.9\std{0.3} & 38.1\std{2.1} & 37.6\std{1.3} & \textbf{41.2\std{0.7}} \\
Final SR (↑) & 23.0\std{2.1} & 24.4\std{2.8} & 23.9\std{1.9} & \textbf{34.2\std{2.1}} \\
\midrule
\textbf{TS-Stellar} & w/o ($\mathbf{f}_{\text{R}} + \mathbf{f}_{\text{S}}$) & w/ $\mathbf{f}_{\text{R}}$ & w/ $\mathbf{f}_{\text{S}}$   & w/ ($\mathbf{f}_{\text{R}} + \mathbf{f}_{\text{S}}$) \\
\midrule
FWT (↑) & \textbf{78.6\std{3.9}} & 76.2\std{4.0} & 73.8\std{1.4} & 75.5\std{0.6} \\
NBT (↓) & 44.0\std{2.1} & 45.0\std{2.6} & 42.7\std{3.1} & \textbf{37.3\std{2.1}} \\
AUC (↑) & 41.9\std{3.1} & 39.1\std{1.6} & 38.1\std{3.5} & \textbf{43.6\std{1.9}} \\
Final SR (↑) & 30.8\std{2.0} & 27.4\std{1.1} & 29.5\std{3.5} & \textbf{35.0\std{3.2}} \\
\bottomrule
\end{tabular}\label{tab:ablation_kroute}
}
\end{minipage}
\end{table*}

\begin{figure}[t]
    \centering
    \includegraphics[width=1\linewidth]{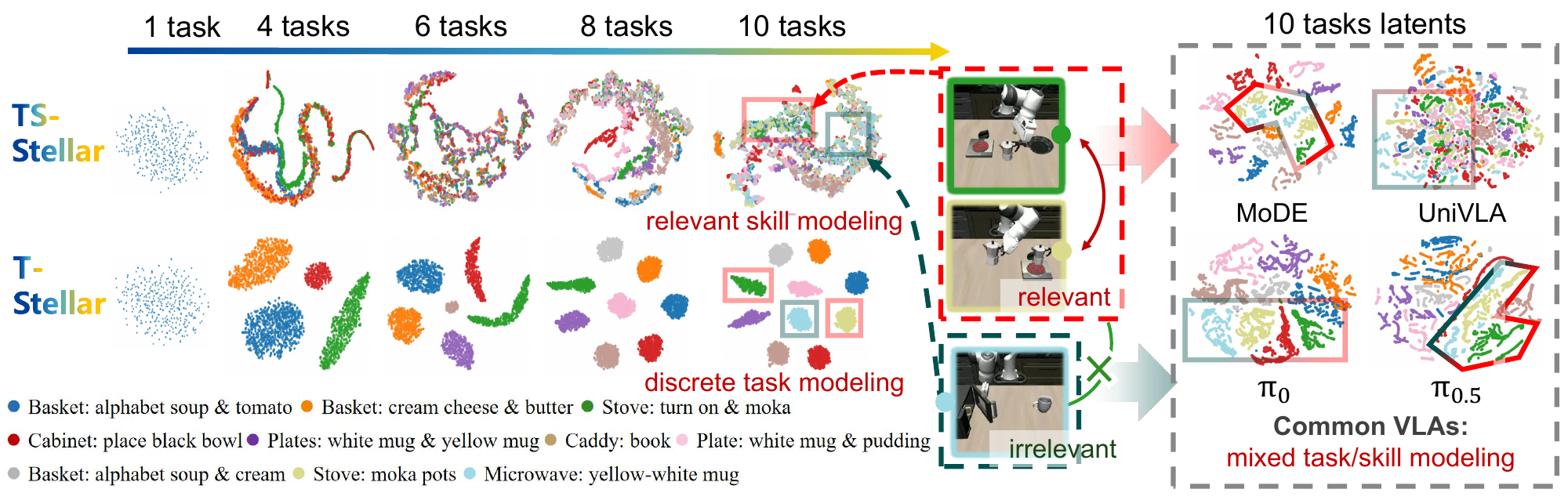}
    \caption{\textbf{T-SNE visualization of Stellar VLA}
Task latents after 1, 4, 6, 8, and 10 tasks on LIBERO-long are shown. Task names are abbreviated for clarity. T-Stellar models discrete task distributions, and TS-Stellar learn relevant skill across tasks, while other VLAs may mix the task modeling.}
    \label{fig:vis}
\end{figure}
\subsection{Ablation Studies}\label{sec:ab}
In this section, we analyze replay efficiency across different budgets, and ablations on the knowledge-space design and expert
routing mechanism of Stellar VLA. Specifically, we compare backbone-matched variants of two key reference methods, MoDE~\cite{MoDE} and LEGION~\cite{legion}, to evaluate the effectiveness of hierarchical task--skill modeling and
knowledge-guided expert routing.

\noindent\textbf{Replay Budget Analysis. } 
Table~\ref{tab:LIBERO_results} demonstrates Stellar VLA's strong performance with only 1\% replay. To further evaluate replay efficiency across different budgets, we compare ER, T-Stellar, and TS-Stellar on LIBERO-long under 1\%, 5\%, and 10\% replay rates (Table~\ref{tab:replay_budget}). While increased replay improves the performance of all models, both Stellar variants consistently outperform ER in FWT, AUC, and Final SR. Notably, Stellar VLA with 5\% replay achieves comparable performance to ER with 10\%, demonstrating greater replay efficiency.

\noindent\textbf{Knowledge Space Ablation. } we replace the Dirichlet-Process-based knowledge space with a standard Gaussian $(\mu=0, \sigma=1)$ (w/o KS) and remove latent task-centric representation learning (w/o VAE). The w/o KS variant is the same as the ER method compared in Section~\ref{sec:main_sim}. The w/o VAE variant serves as a standard MoE baseline similar to MoDE~\cite{MoDE}. Experiments are conducted on all simulation tasks under the scratch setting, as shown in Table ~\ref{tab:ablation_dpmoe}. 
The comparison between w/o VAE and w/o KS shows that task-centric representation learning improves continual learning performance. Our designed knowledge space further provides global task distribution information through $L_{\text{KL}}$ supervision, reducing feature distribution overlap across tasks and mitigating interference between new and previously learned tasks, ultimately improves almost all four metrics.

\noindent\textbf{LEGION-style Conditioning Ablation. } 
As discussed in Section~\ref{sec:intro}, LEGION~\cite{legion} is a key reference for Stellar VLA but targets state-based RL. For direct comparison, we construct a LEGION-style model with task-level DPMM conditioning, using the same vision-language
encoder and action head as Stellar VLA but without posterior-derived relation and semantic routing features. As shown in Table~\ref{tab:legion_style}, both Stellar variants improve AUC and Final SR on LIBERO-goal and LIBERO-long, demonstrating the benefits of knowledge-guided routing and hierarchical task--skill modeling.

\noindent\textbf{Knowledge-Guided Expert Routing Ablation. } We further evaluate Stellar VLA with individual routing features (defined in Section~\ref{sec:moe}) on LIBERO-long, as shown in Table~\ref{tab:ablation_kroute}. The T-Stellar (w/o ($\mathbf{f}_{\text{R}} + \mathbf{f}_{\text{S}}$)) is the same as LEGION-style model in Table~\ref{tab:legion_style}. Both full-routing variants achieve higher AUC and Final SR with complete knowledge guidance. The Knowledge Relation Embedding $\mathbf{f}_{\text{R}}$ captures the relevance between $z$ and existing tasks in the global knowledge space, helping the router decide whether new expert combinations are needed. The Top-K Semantic Embedding $\mathbf{f}_{\text{S}}$ encodes the latent task assignment of $z$, enabling activation of task-relevant expert combinations.

Ablations answer Question (III): Task/skill modeling
structures knowledge to reduce interference
and promote skill reuse, while knowledge-guided routing
selects task-relevant experts. Replay-budget, knowledge-space,
and routing ablations demonstrate their combined effectiveness
in CIL.

\subsection{Continual Task Knowledge Evolution Analysis} \label{sec:vis}
\noindent\textbf{Knowledge Space Visualization. }
To illustrate continual skill evolution in Stellar VLA, we visualize VLA latents on LIBERO-long. Figure~\ref{fig:vis} shows t-SNE plots of latent representations after training on selected tasks. T-Stellar forms distinct task clusters, while TS-Stellar links tasks via shared subskills. 
For instance, the deep and light green clusters overlap due to the shared subgoal “put moka on stove”, while the light blue cluster is separated due to the absence of relevant skills. In contrast, other VLAs~\cite{MoDE, UniVLA, pi0, pi05} either mix these latents or fail to capture task relationship in the feature space.

\begin{wraptable}{r}{0.44\linewidth}
\centering
\vskip -0.1in
\scriptsize
\caption{\textbf{Cross-task subskill similarity.}}
\setlength{\tabcolsep}{2pt}
\renewcommand{\arraystretch}{1.08}
{\fontsize{8.5pt}{10pt}\selectfont\bfseries TS-Stellar subtask similarity}\par\smallskip
\resizebox{0.9\linewidth}{!}{\begin{tabular}{lrr}
\toprule
TV distance $\downarrow$ & Task 3-Sub1 & Task 3-Sub2 \\
\midrule
Task 9-Sub2 & 0.357 & 0.132 \\
Task 9-Sub3 & 0.498 & 0.319 \\
\bottomrule
\end{tabular}}
\par\vspace{6pt}
{\fontsize{8.5pt}{10pt}\selectfont\bfseries T-Stellar subtask similarity}\par\smallskip
\resizebox{0.9\linewidth}{!}{\begin{tabular}{lrr}
\toprule
TV distance $\downarrow$ & Task 3-Sub1 & Task 3-Sub2 \\
\midrule
Task 9-Sub2 & 1.000 & 1.000 \\
Task 9-Sub3 & 1.000 & 1.000 \\
\bottomrule
\end{tabular}}
\label{tab:skill_component_similarity}
\vskip -0.1in
\end{wraptable}

\noindent\textbf{Subskill Similarity Analysis. }To assess whether TS-Stellar's HDP components capture reusable skills, Tasks 3 (``turn on the stove and put the moka pot on it'') and 9 (``put both moka pots on the stove'') from LIBERO-long are selected, with semantic subskill boundaries manually annotated. Task 3-Sub1/Sub2 denote stove activation and first-pot placement, while Task 9-Sub2/Sub3 denote first- and second-pot placement. Subskill similarity is measured by the Total Variation (TV) distance between inferred component-assignment distributions (Table~\ref{tab:skill_component_similarity}), where lower values indicate higher similarity. In TS-Stellar, the shared first-pot placement shows the highest similarity, followed by second-pot placement due to its different target location, while stove activation is least similar to either placement subskill. In contrast, T-Stellar treats cross-task subskills independently due to its task-level modeling. Consistent with Figure~\ref{fig:vis}, these results demonstrate TS-Stellar's ability to discover reusable subskills without explicit trajectory segmentation supervision.

Together, these analyses answer Question (IV): T-Stellar develops task-specific components for new tasks, while TS-Stellar discovers shared task--skill structure across related tasks. The resulting reusable components support continual skill acquisition while preserving prior task knowledge.

\section{Conclusion}
We introduce Stellar VLA, a knowledge space driven policy for continual imitation learning, with two variants: T-Stellar, which constructs task-centric knowledge spaces, and TS-Stellar, which models reusable skills through hierarchical task--skill relations. By using knowledge-posterior features to guide expert selection, our models reduce interference between tasks while enabling continual skill acquisition.
Experiments on LIBERO and real-world tasks demonstrate consistent performance gains over strong baselines.
Stellar VLA provides useful insights into task knowledge understanding, transfer, and utilization in VLA models. Future work could explore larger skill spaces with dynamic hyperparameter adaptation to scale continual learning for larger VLA models.

\newpage
\newpage
\section*{Acknowledgments}
We thank the reviewers for their valuable comments. This work was supported by National Key R\&D Program of China (Grant No.2025YFB4711600). It was also supported in part by the Natural Science Foundation of China under Grant 62225309,U24A20278, 62361166632, the Explorers Program of Shanghai (Basic Research Funding) under Grant 25TS1415100, in part by the Shanghai Municipal Special Program for Basic Research on General AI Foundation Models under Grant 2025SHZDZX025G01.

{\small
\bibliographystyle{unsrt}
\bibliography{ref}
}

\end{document}